\documentclass{article}

\usepackage{microtype}
\usepackage{graphicx}
\usepackage{subfigure}
\usepackage{booktabs} 

\usepackage{amsmath}
\usepackage{amssymb}
\usepackage{nicefrac}
\usepackage{xspace}
\usepackage{multirow}
\usepackage{pifont}

\usepackage[utf8]{inputenc} 
\usepackage[T1]{fontenc}    
\usepackage{microtype}      

\usepackage{hyperref}

\usepackage[accepted]{icml2018}

\graphicspath{{graphics/}}
\newcommand{\Fig}[1]{Figure~\ref{#1}}
\newcommand{\fig}[1]{Fig.~\ref{#1}}
\newcommand{\Tab}[1]{Table~\ref{#1}}
\newcommand{\tab}[1]{Tab.~\ref{#1}}

\newcommand{\eqn}[1]{Eq.~\eqref{#1}} 
\newcommand{\eqnp}[1]{(Eq.~\ref{#1})} 
\renewcommand{\sec}[1]{Section~\ref{#1}} 

\newcommand{\ie}{i.\,e.~}
\newcommand{\eg}{e.\,g.~}
\newcommand{\wrt}{w.\,r.\,t.~}
\newcommand{\Real}{\ensuremath{\mathbb R}}        
\newcommand\Tstrut{\rule{0pt}{2.6ex}}
\DeclareMathOperator*{\argmin}{arg\,min}
\newcommand{\tickNo}{\hspace{1pt}\ding{55}}

\newcommand{\method}{EQL\!$^\div$} 
\newsavebox\MBox
\newcommand\Cline[2][red]{{\sbox\MBox{$#2$}%
  \rlap{\usebox\MBox}\color{#1}\rule[-1.2\dp\MBox]{\wd\MBox}{0.2pt}}}
\newcommand{\mmm}[3]{\ensuremath{#2\,\,^{\scriptscriptstyle #3}_{\Cline[gray]{\scriptscriptstyle #1}}}} 
\newcommand{\layer}[1]{{(#1)}}           
\renewcommand{\l}{{\layer{l}}}           
\newcommand{\lm}{{\layer{l-1}}}           

\newcommand{\citeEQLt}{\citet{MartiusLampert2017:Extrapolation}}

\usepackage{todonotes}

\icmltitlerunning{Learning Equations for Extrapolation and Control}

\begin{document}
\setlength{\abovedisplayskip}{7pt}
\setlength{\belowdisplayskip}{5pt}

\twocolumn[
\icmltitle{Learning Equations for Extrapolation and Control}

\icmlsetsymbol{equal}{*}

\begin{icmlauthorlist}
\icmlauthor{Subham S.~Sahoo}{iit}
\icmlauthor{Christoph H.~Lampert}{ist}
\icmlauthor{Georg Martius}{tu}
\end{icmlauthorlist}

\icmlaffiliation{iit}{Indian Institute of Technology, Kharagpur, India}
\icmlaffiliation{ist}{IST Austria, Klosterneuburg, Austria}
\icmlaffiliation{tu}{Max Planck Institute for Intelligent Systems, T\"ubingen, Germany}

\icmlcorrespondingauthor{Georg Martius}{georg.martius@tuebingen.mpg.de}

\icmlkeywords{deep learning, system identification, discovery, extrapolation, robot control}

\vskip 0.3in
]

\printAffiliationsAndNotice{}  

\begin{abstract}
We present an approach to identify concise equations from data using a shallow neural network approach.
In contrast to ordinary black-box regression, this approach allows understanding functional
relations and generalizing them from observed data to unseen parts of the parameter space.
We show how to extend the class of learnable equations for a recently
proposed equation learning network to include divisions, and we improve
the learning and model selection strategy to be useful for challenging
real-world data.
For systems governed by analytical expressions, our method can in many cases
identify the true underlying equation and extrapolate to unseen domains.
We demonstrate its effectiveness by experiments on a cart-pendulum system,
where only 2 random rollouts are required to learn the forward dynamics
and successfully achieve the swing-up task.
\end{abstract}

\section{Introduction}\label{sec:intro}
In machine learning, models are typically treated as black-box function
approximators that are only judged by their ability to predict correctly
for unseen data (from the same distribution).
In contrast, in the natural sciences, one searches for interpretable
models that provide a deeper understanding of the system of interest
and allow formulating hypotheses about unseen situations.
The latter is only possible if the true underlying functional
relationship behind the data has been identified.
Therefore, when scientists construct models, they do not
only minimize a training error but also impose constraints
based on prior knowledge: models should be \emph{plausible}, \ie
consist of components that have physical expressions in the real
world, and they should be \emph{interpretable}, which typically
means that they consist only of a small number of interacting units.
%

Machine learning research has only very recently started to look into
related techniques. As a first work,~\citeEQLt{}
recently proposed EQL, a neural network architecture for identifying
functional relations between observed inputs and outputs.
Their networks represent only plausible functions through a
specific choice of activation functions and it prefers simple
over complex solutions by imposing sparsity regularization.
However, EQL has two significant shortcomings: first, it is not able
to represent divisions, thereby severely limiting to which physical
systems it can be applied, and second, its model selection procedure
is unreliable in identifying the true functional relation out of
multiple plausible candidates.


In this paper, we propose an improved network for equation learning, \method,
that overcomes the limitation of the earlier works. In particular,
our main contributions are\\[-2em]
\begin{enumerate}\itemsep0em
\item we propose a network architecture that can handle divisions as well as techniques to keep training stable,
\item we improve model/instance selection to be more effective in identifying the right network/equation,
\item we demonstrate how to reliably control a dynamical robotic system by learning its forward dynamics equations from very few random tryouts/tails.
\end{enumerate}

The following section describes the equation learning method by~\citeEQLt{} and
introduces our improvements. Afterwards, we discuss its relation to other
prior work. In \sec{sec:results} we present experimental results on identifying
equations and in \sec{sec:gym} we show its application to robot control.
We close with a discussion and outlook.

\section{Identifying equation with a network}\label{sec:method}
\begin{figure*}
  \centering
  \includegraphics[width=0.6\linewidth]{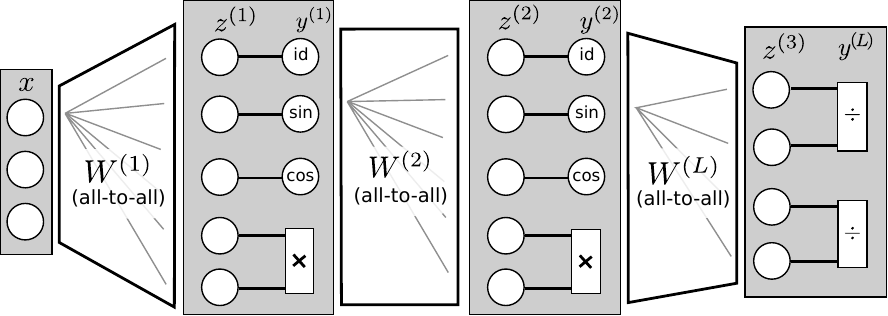}\\[-.6em]
  \caption{Network architecture of the proposed improved Equation Learner \method{} for 3 layers ($L=3$) and one neuron per type ($u=3,v=1$). The new division operations are places in the final layer, see    \citet{MartiusLampert2017:Extrapolation} for the original model.}
  \label{fig:network}
\end{figure*}
We consider a regression problem, where the data originates from a system
that can be described by an (unknown) analytical function, $\phi:\Real^n\to\Real^m$.
A typical example could be a system of ordinary differential equations
that describes the dynamics of a robot, or the predator-prey equations
of an ecosystem.
The observed data, $\{(x_1,y_1),\dots,(x_N,y_N)\}$ 
 is assumed to originate from $y=\phi(x)+\xi$ with additive zero-mean noise $\xi$.
%
Since $\phi$ is unknown, we model the input-output relationship with a function $\psi:\Real^n\to\Real^m$
 and aim to find an instance that minimizes the empirical error on the training set
 as well as on future data, potentially from a different part of the feature space.
For example, we might want to learn the robot dynamics only in a part
of the feature space where we know it is safe to operate, while later
it should be possible also to make predictions for movements into
unvisited areas.


\subsection{Equation Learner}
Before introducing \method, we first recapitulate the working principles
of the previously proposed Equation Learner (EQL) network. It uses a
multi-layer feed-forward network with units representing the building
blocks of algebraic expressions.
Instead of homogeneous hidden units, each unit has a specific function,
\eg identity, cosine or multiplication, see \fig{fig:network}.
Complex functions are implemented by alternating linear
transformations, $z^\l = W^\l y^\lm + w_o^\l$ in layer $l$,
with the application of the base-functions.
There are $u$ unary functions
$f_1,\dots,f_u$, $f_i \in \{\textrm{identity}, \sin, \cos\}$,
which receive the respective component, $z_1,\dots,z_u$.
The $v$ binary functions, $g_1,\dots,g_v$ receive the remaining component, $z_{u+1},\dots,z_{u+2v}$,
as input in pairs of two.
In EQL these are \emph{multiplication units} that compute the product of their two input values: $ g_j(a,b) := a \cdot b$.
The output of the unary and binary units are concatenated to form the output $y^\l$ of layer $l$.
The last layer computes the regression values by a linear read-out
\begin{align}
  y^{\layer{L}} &:= W^{\layer{L}} y^{\layer{L-1}} + w_o^{\layer{L}}.\label{eqn:output}
\end{align}
For a more detailed discussion of the architecture, see~\citep{MartiusLampert2017:Extrapolation}.

\subsection{Introducing division units}
The EQL architecture has some immediate shortcomings. In particular, it
cannot model divisions, which are, however, common in the equations governing physical systems.
%
We, therefore, propose a new architecture,
\method, that includes \emph{division units}, which calculate $\nicefrac{a}{b}$.
Note that this is a non-trivial step because any division creates a
pole at $b\to 0$ with an abrupt change in convexity and diverging function value and its derivative.
Such a divergence is a serious problem for gradient based optimization
methods.

To overcome the divergence problem, we first notice that from any real
system we cannot encounter data at the pole itself because natural
quantities do not diverge.
This implies that a single branch of the hyperbola $\nicefrac{1}{b}$
with $b>0$ suffices as a basis function.
As a further simplification we use divisions only in the ouput layer.
%

Finally, in order to prevent problems during optimization we
introduce a curriculum approach for optimization, progressing from
a strongly regularized version of division to the unregularized
one.

\begin{figure}
  \centering
  \includegraphics[width=0.62\linewidth]{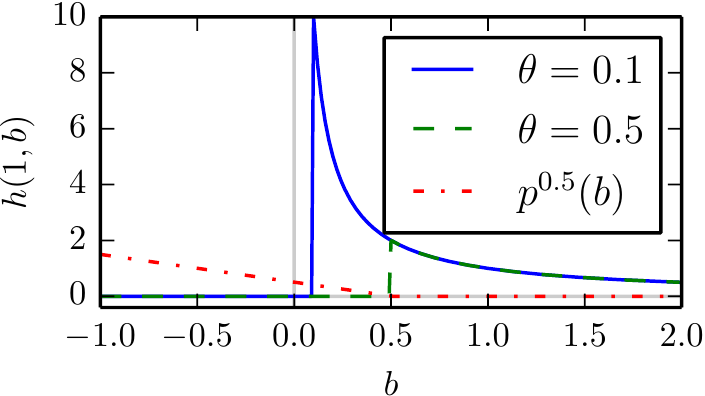}\\[-.8em]
  \caption{Regularized division function $h^\theta(a,b)$ and the associated penalty term $p^\theta(b)$.
    The penalty is linearly increasing for function values $b<\theta$ outside the desired input values.
    \label{fig:div_reg}}
\end{figure}

{\bf Regularized Division:\ }
Instead of EQL's~\eqn{eqn:output}, the last layer of the \method{} is
\begin{align}
  y^{\layer{L}} &:= \Big(h^\theta_1(z^{\layer{L}}_{1},z^{\layer{L}}_{2}),\dots,h^\theta_m(z^{\layer{L}}_{2m},z^{\layer{L}}_{2m+1})  \Big),
  \label{eqn:div_output}
\end{align}
where $h^\theta(a,b)$ is the division-activation function given by
\begin{align}
  h^\theta(a,b) &:= \begin{cases}\frac a b & \text{if } b>\theta\\ 0 &\text{otherwise}
  \end{cases},\label{eqn:div_reg}
\end{align}
where $\theta\geq 0$ is a threshold, see \fig{fig:div_reg}.
Note that using $h^\theta=0$ as the value when the denominator is below $\theta$ (forbidden values of $b$)
sets the gradient to zero, avoiding misleading parameter updates.
So the discontinuity plays no role in practice.


{\bf Penalty term:\ } To steer the network away from negative
values of the denominator, we add a cost term to our objective that
penalizes ``forbidden'' inputs to each division unit:
\begin{align}
  p^\theta(b) &:= \max(\theta - b,0),\label{eqn:penalty}
\end{align}
where $\theta$ is the threshold used in \eqn{eqn:div_reg} and $b$ is the denominator, see \fig{fig:div_reg}.
The global penalty term is then
\begin{align}
P^\theta = \sum_{i=1}^{N}\sum_{j=1}^n p^\theta(z^{\layer{L}}_{2j}(x_i)) \label{eqn:global_penalty}
\end{align}
where  $z^{\layer{L}}_{2j}(x_i)$ is the denominator of division unit $j$ for input $x_i$, see \eqn{eqn:div_output}.

{\bf Penalty Epochs:\ }
While~\eqn{eqn:global_penalty} prevents negative values in
the denominator at training time, the right equation should
not have negative denominators even for potential extrapolation
data.
Similarly, we would like to prevent that output values on future
data have a very different magnitude than the observed outputs,
as this could be a sign of overfitting (\eg learning a polynomial
of too high-degree).
%
To enforce this we introduce particular \emph{penalty epochs},
which are injected at regular intervals (every 50 epochs) into the training process.

During a penalty epoch we randomly sample $N$ input data points
 in the expected test range (including extrapolation region)
 without labels and the network is trained using the cost $\mathcal L^{\textrm{Penalty}} = P^\theta + P^{\text{bound}}$,
 where the latter is given by:
\begin{align}
  P^{\text{bound}} := \sum_{i=1}^{N} \sum_{j=1}^{n} &\max(y^{\layer{L}}_j(x_i)-B,0)\label{eqn:bound} \\
  +&\max(-y_j^{\layer{L}}(x_i)-B,0)\,.\nonumber
\end{align}
The value $B$ reflects the maximal desired output value.
It is, of course, problem dependent but can easily be estimated from
the observed training data. The system is insensitive to the exact choice.
In our experiments, expected outputs are $[-3,3]$ and
we use $B=10$.

{\bf Curriculum:\ } The threshold $\theta$ in the division function \eqn{eqn:div_reg}
 plays the role of avoiding overly large gradients.
However, ultimately we want to learn the precise equation so we
introduce a curriculum during training in which regularization is
reduced continuously. More precisely, $\theta$ decreases with
epoch $t$ as
 \begin{align}
   \theta(t) = 1/\sqrt{t+1}\,.\label{eqn:threshold}
 \end{align}

For validation and testing, we use $\theta=10^{-4}$.

\subsection{Network training}
\method{} is fully differentiable in its free parameters
$\Theta=\{W^{(1)},\dots,W^{(L)},b^{(1)},\dots,b^{(L)}\}$,
which allows us to train it in an end-to-end fashion using back-propagation.
The  objective is Lasso-like~\cite{tibshirani1996regression},
\begin{align}
  \mathcal L&=\!\frac{1}{N}\sum^{N}_{i=1}\|\psi(x_i)-y_i\|^2 + \lambda \sum_{l=1}^L\big|W^\l\big|_1 +P^{\theta}\label{eqn:loss}
\end{align}
that is, a linear combination of $L_2$ loss and $L_1$ regularization extended by the penalty term for small and negative denominators, see \eqn{eqn:penalty}. Note that $ P^{\text{bound}}$ \eqnp{eqn:bound} is only used in the penalty epochs.
For training, we apply a stochastic gradient descent algorithm with mini-batches and Adam~\cite{KingmaBa2015:Adam} for calculating the updates.
The choice of Adam is not critical, as we observed that standard stochastic gradient descent also works, though it might take longer.

{\bf Regularization Phases:\ }
We follow the same regularization scheme as proposed in \citeEQLt{}.
The role of the regularization is to create a network with sparse connections,
 corresponding to the few terms appearing in a typical formula describing a physical system.
Ideally, we would like to minimize the $L_0$ norm directly, but this would make the loss not differentiable.
As commonly done, we use the $L_1$ norm as a substitute.
 However, as detailed below, we add a step for emulating a trade-off free $L_0$ minimization.

\begin{figure}
  \centering
  \includegraphics[width=0.9\linewidth]{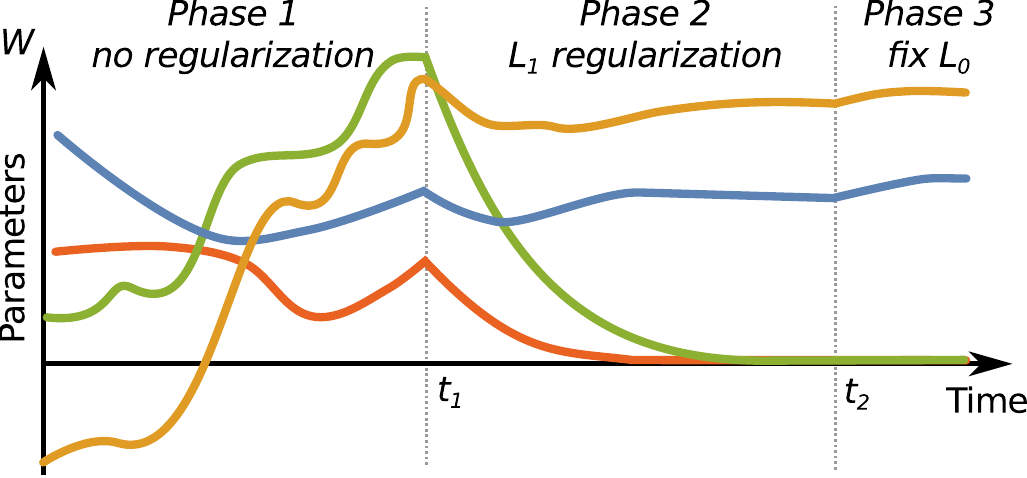}
  \caption{Regularization phases: there is no regularization in the first phase ($t<t_1$)
  where the weights can move freely, followed by a normal $L_1$ phase ($t_1\le t<t_2$) where
  many weights go to zero, followed by a phase ($t_2 \le t$) that fixes the $L_0$ norm by keeping small weights
  at zero and allowing all other weights to go to their correct value. Figure adapted from~\citeEQLt.}
  \label{fig:reg-phases}
\end{figure}

The regularization scheme is as follows:
we start with an un-regularized phase ($\lambda=0$) because starting with $L_1$ regularization sometimes causes weights not to change sign during training but getting ``trapped'' at zero.
We continue by a phase, where regularization is normally enabled by setting $\lambda$ to a nonzero value,
which has the effect that a sparse network structure emerges. Such a phasing was also suggest by~\citet{Carpenter2008:LazyL1Clipping}.
As a side remark, subgradient methods or clipping~\cite{SchmidtFungRosales2009:L1-Opt,Carpenter2008:LazyL1Clipping} was not required to deal with the $L_1$ regularizer because
potentially slight weight fluctuations around zero are not important due to the next phase
which is motivated by the following fact.
A non-zero $L_1$ regularization term leads to a trade-off between minimizing the loss and the regularizer.
Since our aim is to learn the right functional form with the correct coefficients,
 we add a third phase
 where we disable $L_1$ regularization ($\lambda=0$) but enforce the same $L_0$ norm of the weights.
This is achieved by keeping all weights $w\in W^{1\dots L}$ that are close to $0$ at $0$,
 \ie if $|w|<0.001$ then $w=0$ during the remaining epochs.
In this way, the model complexity is fixed and we  ensure
 that function values fit the observed values as closely as possible, allowing for a correct
 parameter estimation. It also eliminates any potential fluctuations of small weights that
 might occur during the $L_1$ phase.
The effect of the regularization phases is schematically illustrated in \fig{fig:reg-phases}.

We use $t_1 = \frac{1}{4} T$ and $t_2=\frac{19}{20} T$, where $T$ is the total number of epochs,
 large enough to ensure convergence, \ie $T=(L-1)\cdot 10000$.
Note, that early stopping will be disadvantageous.

\subsection{Model selection for extrapolation}\label{sec:modelsel}
\newcommand{\ValISparse}{V$^{\text{int}}$-S}
\newcommand{\ValIE}{V$^{\text{int\&ex}}$}
\newcommand{\ValI}{V$^{\text{int}}$}

The model selection is a critical component of the architecture.
Only if the ``right'' formula is selected good extrapolation capabilities can be expected.
A set of different hypothesis equations can be obtained by choosing a range of hyperparameters in particular for the regularization strength.

As suggested in \citeEQLt, the ``right'' formula is singled out by being the simplest one that still predicts well, according to the Occam's razor principle.
We found that this is in principle correct, but for complicated cases we were not able to select the right network instance.
In this paper, we distinguish between two cases, where we have access to a few labeled points in the extrapolation domain and where we do not.

\begin{figure}
  \centering
  \includegraphics[width=0.8\columnwidth]{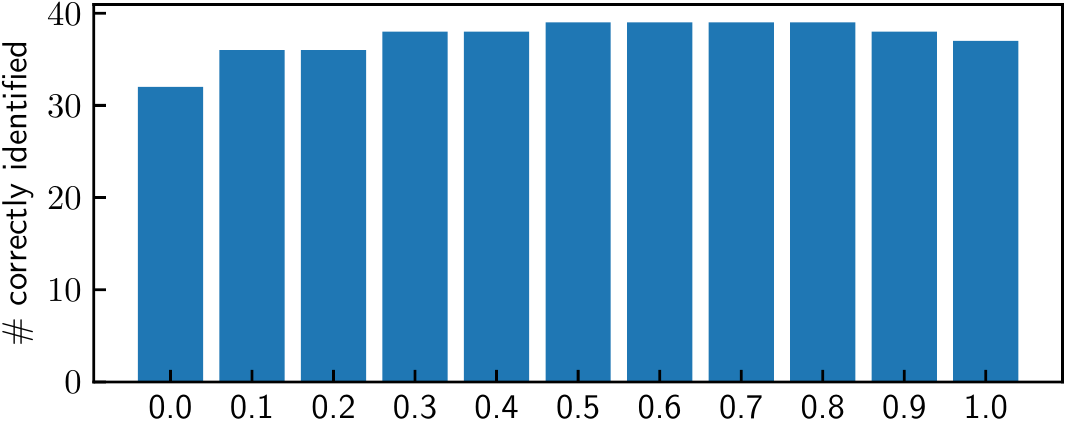}\\[-.6em]
  $\alpha$\\[-1em]
  \caption{Model selection: trade-off between sparsity and validation error, see \eqn{eqn:model:sel} with $\beta=1-\alpha$,
  evaluated empirically in terms of the number of correctly identified equations on all the datasets considered in \sec{sec:results}. We choose the canonical $\alpha=0.5$ as the differences between 0.3 and 0.9 are not significant.}
  \label{fig:modelsel}
\end{figure}

{\bf Without extrapolation data,}
 the selection process has to be based on the validation and sparsity of the instance.
Sparsity is measured in terms of the number of connected units and is denoted by $s$, where a smaller value means simpler/sparser.
In \citet{MartiusLampert2017:Extrapolation}, the best model was selected based on the distance in the space of
 ranked validation error and ranked sparsity.
We found that this method is often disadvantageous because it is insensitive to numeric differences in validation error and practically identically performing instances get different ranks.
In this paper, we instead propose to normalize the quantities instead.
The criterion for the best model is:
 \begin{align}
   \argmin_\psi\left[ \alpha \tilde v^{\textrm{int}}(\psi)^2 +  \beta \tilde s(\psi)^2\right]\,,\label{eqn:model:sel}
 \end{align}
where $\psi$ stands for an instance (trained network), $\tilde v^{\textrm{int}}(\psi)$ and $\tilde s(\psi)$ are
the validation error and sparsity of network $\psi$ normalized to $[0,1]$ \wrt over all instances. The weighting factors $\alpha=0.5$ and $\beta=0.5$ are empirically determined, see \fig{fig:modelsel}.
We call this the {\bf\ValISparse{}} selection method, because it relies on interpolation-validation and sparsity.

{\bf With few extrapolation points,} we form an additional extrapolation-validation dataset
 and denote the validation error as $v^{\text{ex}}$.
In the same manner as above we can select now based on
 \begin{align}
   \argmin_\psi\left[\alpha  \tilde v^{\textrm{int}}(\psi)^2 +  \beta  \tilde s(\psi)^2 + \gamma  \tilde v^{\text{ex}}(\psi)^2\right]\,.\label{eqn:model:sel:2}
 \end{align}
A grid search on the weighting factors $\alpha$, $\beta$, $\gamma$ on all datasets revealed that the sparsity is counterproductive when having some points ($40$ in our case) from the extrapolation domain.
Thus we introduce the {\bf\ValIE{}} selection method using $\alpha=0.5, \beta=0, \gamma=0.5$, \ie same weight on interpolation and extrapolation validation error.
Note that using the normalized values is important because both error terms might be of different scale.

It might be surprising that the sparsity term loses its importance.
We speculate that simply counting nodes can be a misleading measure of simplicity. 
When instead looking at the extrapolation-validation error, the network
can use trigonometric identities with a larger number of terms but which
are easier to find.

How many points do we need from the extrapolation domain?
Since the extrapolation errors vary largely between correct and incorrect formulas, only relatively few
 points are needed. Empirically, around $40$ points were sufficient to identify the right instance from over a hundred candidates.



Because of the strong non-convexity of the problem, the optimization
process may get stuck in a local minimum or not select the correct
formula.
Therefore, to quantify the expected performance deviations, we
use 10 independent runs with random initialization conditions.
\section{Relation to prior work}%

In this work, we are departing from the classical path in machine learning
of finding any function that yields a small expected error on future data
of the same distribution as the training data.
Instead, we aim at discovering the underlying relationship between input and output,
much like it is done in the natural sciences.
In this way an interpretable function with concise form is obtained.

In machine learning, this task has received little attention
 but is studied in the field of \emph{system identification}.
The methods from system identification assume that the functional form of the system is known and only
 the parameters have to be identified.
Recently, this was shown to be effective for identifiying partial differential equations in a range of systems~\cite{RudyKutz2017:PDEDiscovery}.
In our work, we also learn the parameters of the base functions and, most importantly, their composition.

The task of finding equations for observations is also known as symbolic regression.
For a given function space, a heuristic search is performed, typically with evolutionary computation.
With these techniques, it is possible to discover physical laws such as invariants and conserved quantities~\cite{SchmidtLipson2009:learnnaturallaws}.
Unfortunately, due to the exponential search space, the computational complexity becomes prohibitive for larger expressions and high-dimensional problems.
We attempt to circumvent this by modeling it as a gradient-based optimization problem.
Related to symbolic regression is finding mathematical identities for instance to find computationally more efficient expressions. In \cite{ZarembaFergus2014:LearnMathIdentities}, this was done using machine learning to overcome the potentially exponential search space.


%

%

Another relation to our work is in \emph{causal learning},
 which aims at identifying a causal relationship between multiple observables, 
 originating from some underlying functional mechanism.
This was pioneered by \citet{Pearl2000} who reduces this to finding a minimal graphical
model based only on tests of conditional independence.
Although it provides a factorization of the problem and reveals causes
and effects, it leaves the exact functional dependency unexplained.
In order to reveal causal relationships from fewer observables (\eg just two),
 in~\citet{PetersMJS2014} a functional view was taken.
However, the causal inference is based on the expected noise distribution
 instead of the simplicity/plausibility of regression functions.

%
Our work has some similarity to domain adaptation, because data from the
extrapolation domain differs from the data distribution at training
time. As we assume a shared labeling function between both, our
situation fits the covariate shift \cite{quionero2009dataset}
setting. However, existing domain adaptation techniques, such as
sample reweighting \cite{sugiyama2007covariate} are not applicable,
because training and extrapolation domain are disjoint for us. Where
existing approaches rely on an assumption of distribution similarity,
we instead make use of the fact that the target function has an analytic,
and therefore global characterization. For the same reasons, existing
theoretic results, such as \citet{ben2010theory} are not applicable
(or vacuous) in our setting.



For a discussion on the architectural relation to prior work  we refer to \citeEQLt{}.
To summarize, the individual components, such as product units or sine units were introduced before.
However, the combination as introduced here and the restriction to pairwise multiplication terms are new.
To the best of our knowledge, we are the first to use division units in a neural network.

\section{Experimental evaluation}\label{sec:results}
We first demonstrate the ability of \method{} to learn physically inspired
models with and without divisions with good extrapolation quality.
Technically, we implemented the network training and evaluation
procedure in \emph{python} based on the \emph{theano} framework~\cite{2016arXiv160502688short}.
The code and some data is available at \url{https://github.com/martius-lab/EQL}.

For all experiments, we have training data in a restricted domain, usually $[-1,1]^d$ corrupted with noise
 which is split into training and validation with $90\%-10\%$ split.
For testing, we have two sets, one from the training domain and one from an extrapolation domain, for instance $[-2,2]^n\setminus [-1,1]^n$. The error is measured in root mean squares error (RMS) $\sqrt{\frac 1 N \sum_i \|\psi(x_i)-y_i\|^2}$.
The following hyperparameters were fixed: learningrate (Adam) $\alpha=0.001$, regularization (Adam) of $\epsilon=0.0001$, mini-batch size of $20$, number of units $\frac 1 3 u = v = 10$, \ie $10$ units per type in each layer.

\subsection{Learning formulas with divisions}\label{sec:divtask}
We start with a small experiment to check whether simple formulas with divisions can be learned.
We sample data from the following formula:
\begin{align}
y &= \frac{\sin(\pi x_1)}{(x_2^2 + 1)} \label{eqn:div}
\end{align}
As training data, we sample $10000$ points uniformly in the hypercube
{\small $[-1,1]^2$} and add noise ($\sim \mathcal N(0,0.01)$).
The \emph{extrapolation test set} contains $5000$ uniformly sampled points from the data domain
 {\small $[-2,2]^2\setminus [-1,1]^2$}.
Note, that the extrapolation domain is 3 times larger than the training domain.
We perform model selection among the following hyper-parameters:
the regularization strength {\small $\lambda\in 10^{\{-6,-5.9,\dots,-3.6,-3.5\}}$},
 and {\small $L\in\{2,3,4\}$}.
We use the same parameters for all experiments.
We compare our algorithm to a standard multilayer perceptron (MLP) with $\tanh$ activation functions and
possible hyperparameters: $\lambda\in 10^{\{-6.3, -6.0, -5.3, -5.0, -4.3, -4.0\}}$,
 number of layers {\small $L\in\{2,3,4\}$}, and number of neurons {\small $k\in\{5,10,20\}$}.
A second baseline is given by epsilon support vector regression (SVR)~\cite{basak2007:SVR} with
 two hyperparameters {\small $C\in10^{\{-3,-2,-1,0,1,2,3,3.5\}}$} and {\small $\epsilon \in 10^{\{-3,-2,-1,0\}}$}
 using radial basis function kernel with width {\small $\gamma\in \{0.05,0.1,0.2,0.5,1.0\}$}.
We also compare to Eureqa~\cite{eureqa:website}, a symbolic regression algorithm using evolutionary search introduced in \cite{SchmidtLipson2009:learnnaturallaws}. The termination condition was when the software reported $100\%$ convergence and no better solution was found in the last $10\%$ of the search time.

\begin{figure}
  \centering
  \begin{minipage}[t]{0.4\linewidth} \centering \small
    (a)\\[-1em]
    \begin{tabular}[t]{l|l}
      \toprule
      & extrapol.\\
      \midrule
      \method & \mmm{0.01}{0.01}{0.01}\\
      EQL     & \mmm{0.07}{0.20}{0.26}\\
      MLP     & \mmm{0.77}{0.83}{1.00}\\
      SVR     & $0.26$\\
      Eureqa  & \mmm{0.01}{0.13}{0.15}\\
      \bottomrule
    \end{tabular}
  \end{minipage}%
  \begin{minipage}[t]{0.59\linewidth}\centering
    (b)\\
    \includegraphics[width=\linewidth]{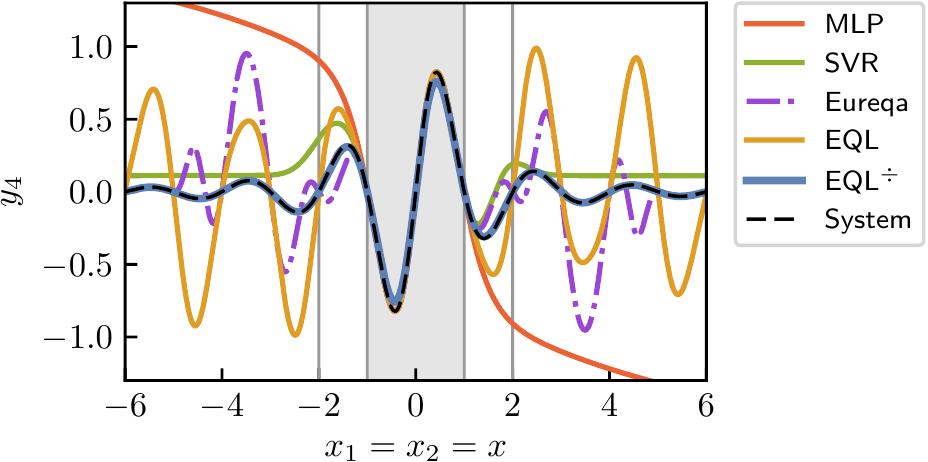}
  \end{minipage}
  \vspace*{-6pt}
  \caption{Results on the \emph{division} task, \eqn{eqn:div}.
    (a) Reported are median, minimum and maximum (in sub and superscript) of the root mean squares error (RMS) for $10$ random initializations. (b) Output $y$ for a slice of the inputs $x_1=x_2=x$
    for the true system equation \eqnp{eqn:div}, and best model-selected instances (for the run with the median performance).
    Note that we plot a much larger domain than that of the extrapolation test-set.
  }\label{fig:div}
\end{figure}

In \fig{fig:div} the numerical results and also an illustrative output of \method{} and the baselines are presented.
Only \method{} can extrapolate to the unseen domain because it has the capacity to identify the underlying expression, which was achieved in all of the $10$ independent runs.
Note, that the original EQL method cannot extrapolate because it lacks the division units.
Interestingly, Eureqa is not finding the right equations in $4/5$ cases.
The interpolation performances are at the noise level of $0.01$ for all methods (not shown).

\subsection{Learning complex formulas.}\label{sec:syn}

Following \citeEQLt{} we test \method{} on a set of complicated formulas and compare it to
 the original EQL, MLP, SVR and Eureqa as baselines. These do not contain divisions and we want to test whether
 this poses any problems as our architecture always contains divisions.
We consider the following formulas with four-dimensional input and one-dimensional output:
{
\begin{align*}
y &= \nicefrac{1}{3} \left(\sin(\pi x_1) + \sin\left(2 \pi x_2 + \nicefrac{\pi}{8}\right)+x_2 - x_3 x_4 \right)&\text{(F-1)}\\
y &= \nicefrac{1}{3} \left(\sin(\pi x_1)\!+\!x_2 \cos(2\pi x_1\!+\!\nicefrac{\pi}{4})\!+\!x_3\!-\!x_4^2\right) &\text{(F-2)}\\
y &= \nicefrac{1}{3} \left( (1+x_2)  \sin(\pi x_1) + x_2  x_3  x_4\right) &\text{(F-3)}\\
y &= \nicefrac{1}{2} \left(\sin(\pi x_1) + \cos(2 x_2 \sin(\pi x_1))+ x_2  x_3 x_4\right) &\text{(F-4)}
\end{align*}
}
For a correct identification, the equations F-1 requires one hidden layer, F-2 and F-3 require two, and F-4 requires three ($L=4$).
Nevertheless, the right number of layers is automatically detected by model selection.
%
Data generation and training procedure is the same as for the \sec{sec:divtask},
 except that the data is now from a 4-dimensional hypercube, which makes the
extrapolation domain 15 times larger than the training domain.

\Tab{tab:syn:results} shows the numerical results.
All methods are able to interpolate, but only \method{} and Eureqa achieve
good extrapolation results in all cases.
The original EQL (with the original model selection) was able to find the right answer only for F-2.
With the model selection \ValIE{} (using 40 extrapolation points), \method{} and Eureqa are able to
 find the right formula for F-3 every time and for F-4 in about half of the cases (manual inspection of the learned equations).

\begin{figure}
  \centering
  \begin{tabular}{c@{\ }c}
    (a) F-4 &(b) RE2-2\\
    \includegraphics[height=0.25\linewidth]{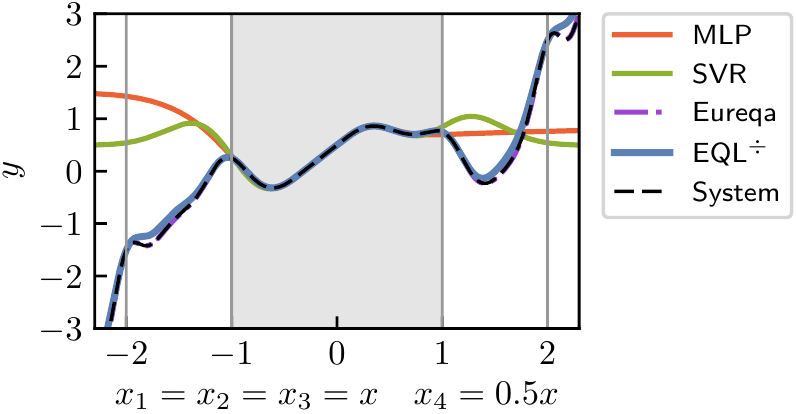}
    &\includegraphics[height=0.25\linewidth]{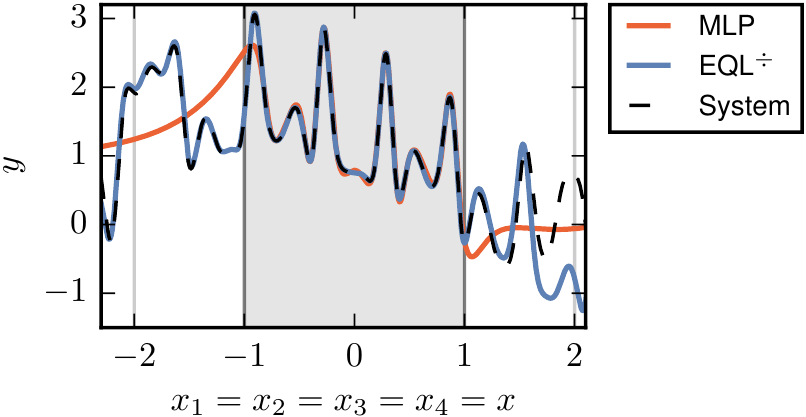}
  \end{tabular}
  \vspace*{-10pt}
  \caption{\emph{Formula learning}: (a) for F-4, (b) for RE3-4.
    Shown is $y$ for a single cut through the input space for the true system equation
    and for the best model-selected instances. 
  }\label{fig:syn:slice}
\end{figure}

\Fig{fig:syn:slice}(a) illustrates the performance of the learned models for F-4 visually.
It shows the output for a slice through the input for one of the model-selected instances for each method.
It is remarkable how well the extrapolation works for \method{} and Eureqa even far outside training region.

\begin{table}
  \centering
  \begin{tabular}{ll|ll}
    \toprule
    &&\multicolumn{2}{c}{extrapolation error}\\
    dataset&method& \ValISparse/\ValI & \ValIE \\\hline
    F-2\Tstrut
    & \method{}  & \mmm{0.01}{0.01}{0.01} & \mmm{0.01}{0.01}{0.01} \\
    & EQL        & \mmm{0.01}{0.03}{0.06} & --- \\ 
    & \text{MLP} & \mmm{0.50}{0.58}{0.64} & \mmm{0.43}{0.46}{0.50} \\
    & \text{SVR} & $0.91         $ & $0.45        $ \\
    & \text{Eureqa} & \mmm{0.01}{0.01}{0.01} & \mmm{0.01}{0.01}{0.01} \\
    \hline
    F-3\Tstrut
    &\method     & \mmm{0.01}{0.01}{0.01} & \mmm{0.01}{0.01}{0.01}\\
    &EQL         & \mmm{0.15}{0.35}{0.51} & --- \\ 
    &\text{MLP}  & \mmm{0.43}{0.47}{0.51} & \mmm{0.39}{0.46}{0.50} \\
    &\text{SVR}  & $0.34$          & $0.39$ \\
    & \text{Eureqa} & \mmm{0.01}{0.01}{0.33} & \mmm{0.01}{0.01}{0.46} \\
    \hline
    F-4\Tstrut
    &\method     & \mmm{0.07}{0.23}{1.04} & \mmm{0.07}{0.21}{0.71} \\
    &EQL         & \mmm{0.25}{0.37}{0.46} & --- \\ 
    &\text{MLP}  & \mmm{0.72}{0.86}{0.95} & \mmm{0.72}{0.86}{0.95} \\
    &\text{SVR}  & $0.91$         & $1.28$ \\
    & \text{Eureqa} & \mmm{0.54}{0.85}{0.95} & \mmm{0.01}{0.18}{1.10} \\
    \bottomrule
  \end{tabular}
  \caption{Extrapolation performance and model selection for \emph{formula learning}. See \fig{fig:div} for details. For EQL the original model selection was used. For F-1 (not shown), both EQL and \method{}, have an error of 0.01. }\label{tab:syn:results}
\end{table}

\subsection{Random expressions}\label{sec:rg} 
In order to avoid a bias through hand-crafted formulas, we generated random functional expressions.
The expressions were generated with our architecture with $2$ and $3$ hidden layers and
  random sparse connections, $4$ instances each, named as RE$\{2,3\}$-$\{1,2,3,4\}$.
The weights and biases were sampled from a uniform distribution in $[0.5, 2]$ and were multiplied with $[1,-1]$
with equal probability. Further, the input weights into the $\sin$ and $\cos$ nodes were multiplied by $\pi$,
to avoid including only the linear regime of the trigonometric functions.
%
Some of the expressions are very hard to learn, even in the interpolation region.
\Tab{tab:rg:results} shows the experimental results.
We also compare the two model-selection strategies (\ValISparse{} and \ValIE{}).
It becomes evident that without a few points from the extrapolation region (here $40$) the system is not able
 to identify the right formula in the majority of cases.
For the baselines (MLP, SVR) the model selection based on extrapolation only reduced the strong outliers
 but did not yield acceptable performance. This is expected because these methods have no chance
 to identify the right functional relationship.
\Fig{fig:syn:slice}(b) illustrates the complicated structure at the example of RE3-4.
Again, note that the extrapolation domain is 15 times larger than the training domain.

\begin{table*}
\caption{Extrapolation performance for \emph{random graphs}.
  See \fig{fig:div} for details. Results for different methods and model selection schemes.
  Const 0 refers to a constant prediction of zero. For some random expressions marked with \tickNo (RE2-3 and R3-1) we are not able to learn them with satisfactory precision. A visualization of RE3-4 can be found in \fig{fig:syn:slice}(b).
}\label{tab:rg:results}
\centering
\begin{tabular}{l@{\ }l@{ }|llllllll@{ }}
  \toprule
  && RE2-1 & RE2-2 & RE2-3 \tickNo & RE2-4 & RE3-1 \tickNo & RE3-2 & RE3-3 & RE3-4\\
  \hline \Tstrut

  \method  & \ValIE{}  & $\mmm{0.02}{0.02}{0.02}$ & $\mmm{0.03}{0.04}{0.11}$ & $\mmm{0.48}{0.52}{0.82}$ & $\mmm{0.01}{0.01}{0.01}$ & $\mmm{0.28}{0.46}{0.55}$ & $\mmm{0.01}{0.02}{0.06}$ & $\mmm{0.01}{0.01}{0.01}$ & $\mmm{0.02}{0.03}{0.52}$\\
  \method  & \ValISparse{} & $\mmm{0.02}{0.27}{0.39}$ & $\mmm{0.14}{0.14}{0.14}$ & $\mmm{0.55}{0.76}{2.05}$ & $\mmm{0.01}{0.01}{0.01}$ & $\mmm{0.31}{0.51}{1.23}$ & $\mmm{0.04}{0.08}{4.65}$ & $\mmm{0.01}{0.01}{0.01}$ & $\mmm{0.02}{0.03}{1.64}$\\
  MLP      & \ValIE{}  & $\mmm{1.43}{1.54}{1.66}$ & $\mmm{0.96}{1.04}{1.09}$ & $\mmm{0.87}{0.90}{0.91}$ & $\mmm{0.86}{0.95}{1.12}$ & $\mmm{0.84}{1.04}{1.36}$ & $\mmm{1.60}{1.85}{2.13}$ & $\mmm{0.40}{0.52}{0.58}$ & $\mmm{1.34}{1.64}{1.96}$\\
  MLP      & \ValI{}  & $\mmm{1.44}{1.60}{1.66}$ & $\mmm{1.01}{1.05}{1.10}$ & $\mmm{1.10}{1.47}{1.65}$ & $\mmm{0.86}{0.99}{1.16}$ & $\mmm{1.07}{1.31}{1.59}$ & $\mmm{1.65}{2.03}{2.24}$ & $\mmm{0.73}{1.16}{2.02}$ & $\mmm{1.61}{1.89}{2.12}$\\
  SVR      & \ValIE{}  & $1.15$ & $1.09$ & $0.59$ & $1.51$ & $0.96$ & $1.81$ & $0.37$ & $1.23$\\
  SVR      & \ValI{}  &$1.20$ & $2.12$ &  $17.72$ & $13.89$ & $11.79$ & $11.28$ & $0.37$ & $17.67$ \\
  \hline
  Const & 0         & $6.73$ & $2.57$ & $0.50$ & $5.36$ & $1.65$ & $72.26$ & $17.67$ & $3.15$\\
 \bottomrule
\end{tabular}
\end{table*}

\subsection{Cart-pendulum system}
\begin{figure}
  \centering
  \begin{tabular}{c@{\ \ }c}
    (a) &(b)\\
      \includegraphics[width=0.4\linewidth]{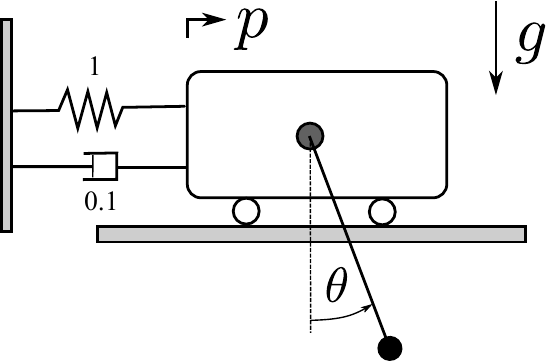}&
    \includegraphics[width=0.55\linewidth]{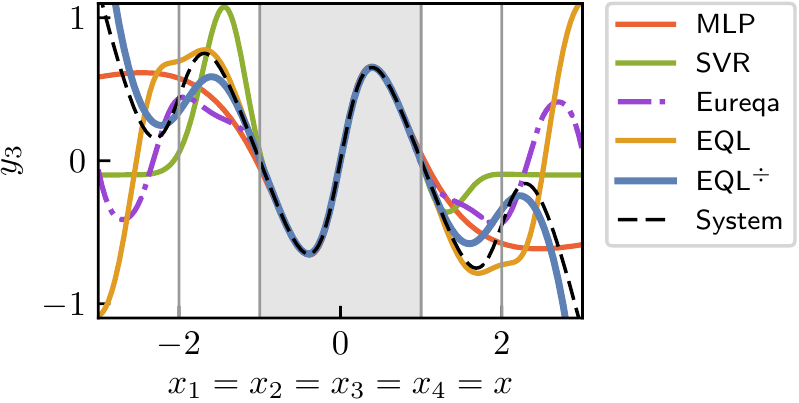}
  \end{tabular}
  \vspace*{-10pt}
  \caption{Cart-pendulum system. (a) sketch of the system.
     The lengths and masses are set to 1, the gravitation constant is $9.81$ and the friction constant is $0.01$.
    (b) slices of outputs $y_3$ for inputs $x_1=x_2=x_3=x_4=x$
     for the true system equation \eqnp{eqn:cp}, and learned EQL, \method{}, and MLP instances.
   }
  \label{fig:cp}
\end{figure}

Let us now consider a non-trivial physical system:
 a pendulum attached to a cart that can move horizontally along a rail but that is attached to a
spring damper system, see \fig{fig:cp}(a).
The system is parametrized by $4$ unknowns: the position of the cart,
the velocity of the cart, the angle of the pendulum and the angular
velocity of the pendulum. We combine these into a four-dimensional
vector $x=(x_1,\dots,x_4)$.
We set up a regression problem with four outputs from the corresponding
system of ordinary differential equations where $y_1 = \dot x_1 = x_3$, $y_2 = \dot x_2 = x_4$ and
\begingroup\makeatletter\def\f@size{5}\check@mathfonts
\def\maketag@@@#1{\hbox{\m@th\large\normalfont#1}}%
\begin{align}
  y_3&=\!\frac{-x_1-0.01 x_3+x_4^2 \sin\left(x_2\right)+0.1 x_4
    \cos \left(x_2\right)\!+\!9.81 \sin \left(x_2\right) \cos
    \left(x_2\right)}{\sin ^2\left(x_2\right)+1}\label{eqn:cp}\\[-.1em]
  y_4&=\!\frac{-0.2 x_4\!-\!19.62 \sin\!\left(x_2\right)\!+\!x_1\!
    \cos\!\left(x_2\right)\!+\!0.01 x_3\!\cos\! \left(x_2\right)\!-\!x_4^2
    \sin\!\left(x_2\right)\!\cos\!\left(x_2\right)}
  {\sin^2\left(x_2\right)+1}.\nonumber
\end{align}
\endgroup
The task is to learn the function without controlling the system.
The formulas contain divisions which are now included in the \method{} architecture.
In \fig{fig:cp}(b) the extrapolation performance is illustrated by slicing through the input space.
Near the training region all methods fit the data well, but a bit further away
 only \method{} can predict well.
For all other methods, even the best instances
 differ considerably from the true values, see also the numerical results in \tab{tab:cp:results}.
In 1 out of the 10 independent runs also \method performed poorly. This is less likely for a finer scan of $\lambda$ values.


\begin{table}
\caption{Interpolation and extrapolation performance for \emph{cart-pendulum dynamics}.
  See \fig{fig:div} for details. \ValIE{} model selection is used.
  Note that predicting 0 would yield an error of 0.96.}\label{tab:cp:results}
\centering

\begin{tabular}{l|ll}
  \toprule
   & \text{interpolation} & \text{extrapolation} \\
  \hline \Tstrut
  \method      & \mmm{0.010}{0.010}{0.010}  & \mmm{0.04}{0.06}{4.20} \\
   EQL         & \mmm{0.010}{0.010}{0.011}  & \mmm{0.16}{0.17}{0.21} \\ 
  \text{MLP}   & \mmm{0.012}{0.012}{0.012}  & \mmm{0.18}{0.18}{0.19} \\
  \text{SVR}   & $0.019$                  & $0.36         $ \\
  \text{Eureqa}& \mmm{0.011}{0.012}{0.012} & \mmm{0.15}{0.19}{17.2}\\
 \bottomrule
\end{tabular}
\end{table}

\section{Control using learned dynamics}\label{sec:gym}
In this section, we will demonstrate the effectiveness of the equation learning for robot control.
For the cart-pole we know that the system can learn correctly the differential equations from randomly sampled data, see above.
Now, we are challenging the system to learn the dynamics from actual interactions and subsequently use it
 to control the robot, namely to learn how to perform the swing-up task with the cart-pole.
We use the OpenAI Gym cart-pole environment~\cite{aigym:cartpole} that we modified for the swing-up task, \ie to start at the bottom.
The state of the system $s=(x,\dot x, \theta, \dot \theta)$ contains the position and velocity of the cart and the angle and angular velocity of the pole.
The learner should model the forward dynamics $f(s,a) \to \dot s$, predicting state changes from the current state and action.

At the beginning no knowledge about the system is available, such that we perform $K-1$ random rollouts with $1000$ steps ($20$secs)
 with random actions $a\sim\mathcal N(0,0.15)$.
The resulting pole angle distribution had mean $\pi$ and standard deviation of approximately $\nicefrac{\pi}{4}$, so only a small part of the angle range was visited.
In order to obtain data for model selection we perform one additional rollout with $a\sim\mathcal N(0,0.25)$.

After training we use the resulting models for optimal control.
We define a cost which defines the desired position, \ie
 vertical upright pole with small velocities and cart at the center:
\begin{align}
  C = 0.1 x^2 + 0.1\dot x^2 - \cos(\theta) + 0.02\dot \theta^2\label{eqn:mpccost}\,.
\end{align}
Note, that the stability point lies far outside the training domain.
To achieve the optimal control we use model predictive control (MPC)~\cite{GarciaPrettMorari1998:MPC}
 with a random shooting method.
Briefly, every timestep 1000 lookaheads are simulated using the learned dynamics (Euler integration)
 with random actions. After a fixed horizon the look-ahead with the lowest cost $C$ is chosen and
 only the first action is applied. This runs in realtime for $100$\,Hz update rate.
For MPC to work the model should provide with a good enough representation of the system dynamics.
 \begin{figure}
   \centering
   \begin{tabular}{c@{\ \ }c}
     (a) reward vs.~number of rollouts $K$&(b) noisy system\\
     \includegraphics[height=.4\columnwidth]{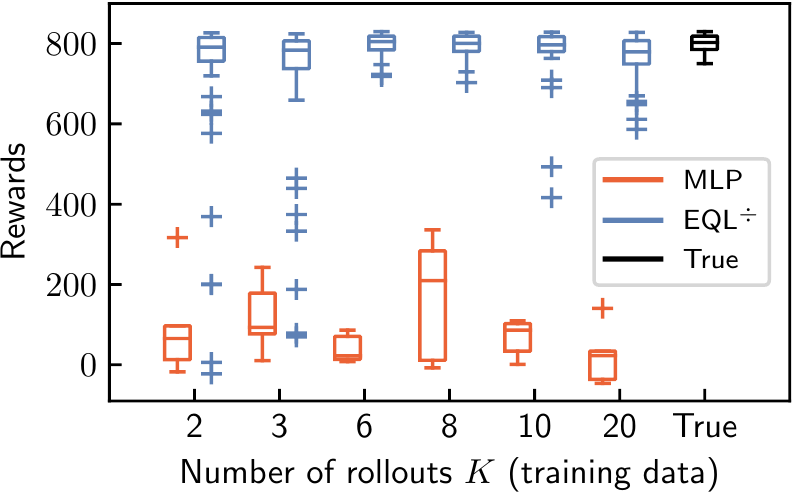}&%
     \includegraphics[height=.4\columnwidth,trim=0 -3pt 0 0]{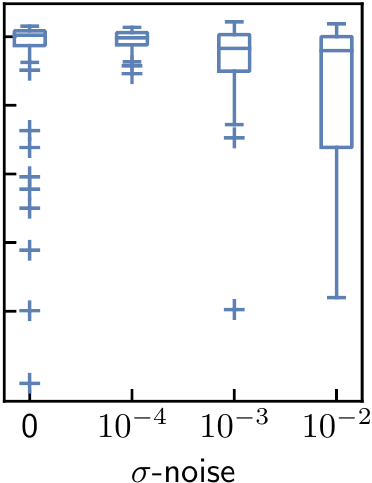}%
   \end{tabular}
   \vspace*{-1.2em}
   \caption{Performance on the cart-pole swing-up task with learned dynamics.
     (a) reward during test-runs for models trained on $K$ rollouts (5 independent experiments with 10 test runs each).
     (b) robustness to noise in sensors and actions for $K=3$.
   }
   \label{fig:cpmpc}
 \end{figure}

In \fig{fig:cpmpc} the results for different number of rollouts $K$ are presented.
The reward from the environment is $R=\sum_t \cos(\theta_t)$.
We report statistics over $5$ independent experiments each.
With \method{} the cart-pole is able to accomplish the task
 already after $K=2$ rollouts (one for training, one for validation) in most of the cases.
For $K=3$ we also show its robustness to noise on the states and actions.
In comparison the neural network forward model (MLP) does only perform swing around operation
 but was nowhere close to stabilizing it in the vertical position.

\begin{figure}
   \centering
   \includegraphics[width=.48\columnwidth]{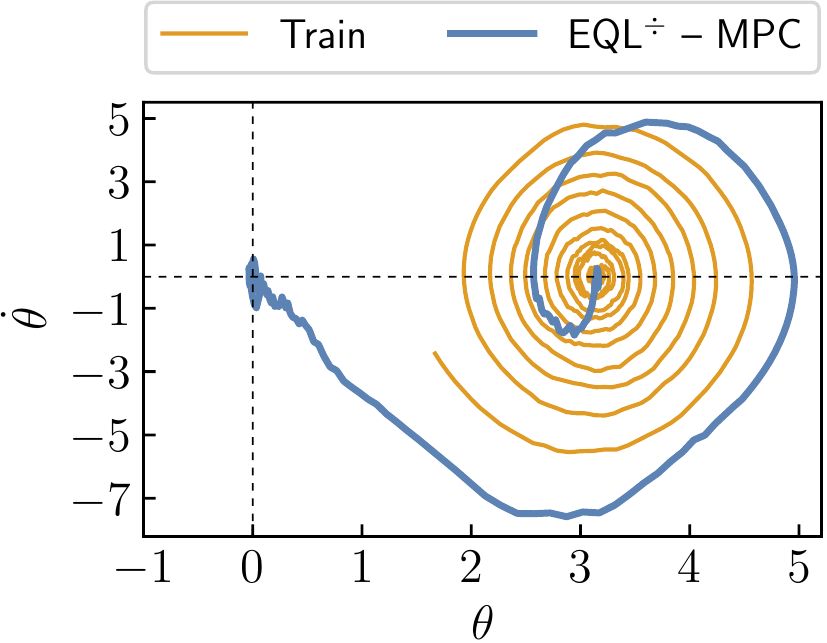}%
   \includegraphics[width=.48\columnwidth, trim=0 -1pt 0 0]{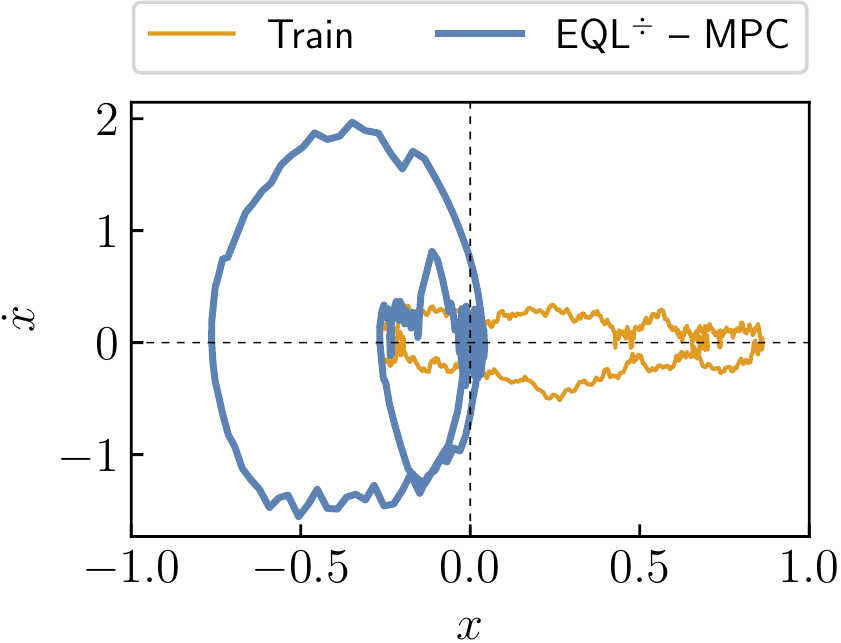}
   \vspace*{-1em}
   \caption{Training trajectory and controlled trajectory (\method) of cart-pole system in angle-space left and cart positions right, $K=2$. See also the Video at \url{https://youtu.be/MG9q3gTtBLs}.
   }
   \label{fig:cp-traj}
 \end{figure}

In \fig{fig:cp-traj} a sample training trajectory and a final control trajectory is displayed for $K=2$.

\section{Conclusions}
In this paper, we introduce \method, a new network architecture for
equation learning. It improves over previously suggested work by
including the ability to learn division, thereby substantially
increasing the applicability.
%
The network is fully differentiable and trainable with back-propagation,
which we achieve by introducing a regularized version of division that
 is smoothly transformed towards the true division in a curriculum fashion.
We also introduce a new model selection technique that identifies
the actual functional relation between inputs and outputs more
reliably than the purely sparsity-based approach of previous work.


We empirically demonstrate that \method{} can learn various functional
relations, with and without divisions, from noisy data in a confined domain.
Furthermore, those can extrapolate to unseen parts of the data space.
In a broad set of experiments we show that the approach learns concise
functional forms that can be inspected and may provide insights into
the relationships
within the data. 
%


This ability opens doors for many new applications. As an exemplary task,
we show efficient model learning for robot control: \method{} identifies
the forward dynamics for an actuated cart-pendulum system from just
2 random rollouts with limited excitation and under noisy observations.
The resulting forward model is good enough to robustly perform a
pendulum swing-up task, despite the fact that such a pattern was never
observed during training.
Also embedding \method{} as part of deep architecture is conceivable.
In future studies, we will consider larger and more complex systems, where we expect good scaling capabilities due to the gradient-based optimization.




\begin{thebibliography}{18}
\providecommand{\natexlab}[1]{#1}
\providecommand{\url}[1]{\texttt{#1}}
\expandafter\ifx\csname urlstyle\endcsname\relax
  \providecommand{\doi}[1]{doi: #1}\else
  \providecommand{\doi}{doi: \begingroup \urlstyle{rm}\Url}\fi

\bibitem[Basak et~al.(2007)Basak, Pal, and Patranabis]{basak2007:SVR}
Basak, D., Pal, S., and Patranabis, D.~C.
\newblock Support vector regression.
\newblock \emph{Neural Information Processing-Letters and Reviews}, 11\penalty0
  (10):\penalty0 203--224, 2007.

\bibitem[Ben-David et~al.(2010)Ben-David, Blitzer, Crammer, Kulesza, Pereira,
  and Vaughan]{ben2010theory}
Ben-David, S., Blitzer, J., Crammer, K., Kulesza, A., Pereira, F., and Vaughan,
  J.~W.
\newblock A theory of learning from different domains.
\newblock \emph{Machine Learning}, 79\penalty0 (1-2):\penalty0 151--175, 2010.

\bibitem[Carpenter(2008)]{Carpenter2008:LazyL1Clipping}
Carpenter, B.
\newblock Lazy sparse stochastic gradient descent for regularized multinomial
  logistic regression.
\newblock \emph{Alias-i, Inc., Tech. Rep}, pp.\  1--20, 2008.

\bibitem[García et~al.(1989)García, Prett, and
  Morari]{GarciaPrettMorari1998:MPC}
García, C.~E., Prett, D.~M., and Morari, M.
\newblock Model predictive control: Theory and practice—a survey.
\newblock \emph{Automatica}, 25\penalty0 (3):\penalty0 335 -- 348, 1989.
\newblock ISSN 0005-1098.
\newblock \doi{10.1016/0005-1098(89)90002-2}.

\bibitem[Kingma \& Ba(2015)Kingma and Ba]{KingmaBa2015:Adam}
Kingma, D.~P. and Ba, J.
\newblock Adam: A method for stochastic optimization.
\newblock In \emph{Proceedings of ICLR}, 2015.

\bibitem[Martius \& Lampert(2016)Martius and
  Lampert]{MartiusLampert2017:Extrapolation}
Martius, G. and Lampert, C.~H.
\newblock Extrapolation and learning equations, 2016.
\newblock arXiv preprint \url{https://arxiv.org/abs/1610.02995}.

\bibitem[Nutonian(2018)]{eureqa:website}
Nutonian.
\newblock Eureqa desktop software.
\newblock \url{https://www.nutonian.com/products/eureqa-desktop}, 2018.

\bibitem[{OpenAI Gym}(2018)]{aigym:cartpole}
{OpenAI Gym}.
\newblock Cartpole-v0.
\newblock \url{https://gym.openai.com/envs/CartPole-v0}, 2018.

\bibitem[Pearl(2000)]{Pearl2000}
Pearl, J.
\newblock \emph{Causality}.
\newblock Cambridge {U}niversity {P}ress, 2000.

\bibitem[Peters et~al.(2014)Peters, Mooij, Janzing, and
  Sch{\"o}lkopf]{PetersMJS2014}
Peters, J., Mooij, J., Janzing, D., and Sch{\"o}lkopf, B.
\newblock Causal discovery with continuous additive noise models.
\newblock \emph{Journal of Machine Learning Research (JMLR)}, 15:\penalty0
  2009--2053, 2014.

\bibitem[Quionero-Candela et~al.(2009)Quionero-Candela, Sugiyama, Schwaighofer,
  and Lawrence]{quionero2009dataset}
Quionero-Candela, J., Sugiyama, M., Schwaighofer, A., and Lawrence, N.~D.
\newblock \emph{Dataset shift in machine learning}.
\newblock The MIT Press, 2009.

\bibitem[Rudy et~al.(2017)Rudy, Brunton, Proctor, and
  Kutz]{RudyKutz2017:PDEDiscovery}
Rudy, S.~H., Brunton, S.~L., Proctor, J.~L., and Kutz, J.~N.
\newblock Data-driven discovery of partial differential equations.
\newblock \emph{Science Advances}, 3\penalty0 (4), 2017.
\newblock \doi{10.1126/sciadv.1602614}.

\bibitem[Schmidt \& Lipson(2009)Schmidt and
  Lipson]{SchmidtLipson2009:learnnaturallaws}
Schmidt, M. and Lipson, H.
\newblock Distilling free-form natural laws from experimental data.
\newblock \emph{Science}, 324\penalty0 (5923):\penalty0 81--85, 2009.
\newblock ISSN 0036-8075.
\newblock \doi{10.1126/science.1165893}.

\bibitem[Schmidt et~al.(2009)Schmidt, Fung, and
  Rosales]{SchmidtFungRosales2009:L1-Opt}
Schmidt, M., Fung, G., and Rosales, R.
\newblock Optimization methods for {$L_1$}-regularization.
\newblock UBC Technical Report TR-2009-19, 2009.

\bibitem[Sugiyama et~al.(2007)Sugiyama, Krauledat, and
  Müller]{sugiyama2007covariate}
Sugiyama, M., Krauledat, M., and Müller, K.-R.
\newblock Covariate shift adaptation by importance weighted cross validation.
\newblock \emph{Journal of Machine Learning Research}, 8\penalty0
  (May):\penalty0 985--1005, 2007.

\bibitem[{Theano Development Team}(2016)]{2016arXiv160502688short}
{Theano Development Team}.
\newblock {Theano: A {Python} framework for fast computation of mathematical
  expressions}.
\newblock \emph{arXiv e-prints}, abs/1605.02688, May 2016.
\newblock URL \url{http://arxiv.org/abs/1605.02688}.

\bibitem[Tibshirani(1996)]{tibshirani1996regression}
Tibshirani, R.
\newblock Regression shrinkage and selection via the lasso.
\newblock \emph{Journal of the Royal Statistical Society. Series B
  (Methodological)}, pp.\  267--288, 1996.

\bibitem[Zaremba et~al.(2014)Zaremba, Kurach, and
  Fergus]{ZarembaFergus2014:LearnMathIdentities}
Zaremba, W., Kurach, K., and Fergus, R.
\newblock Learning to discover efficient mathematical identities.
\newblock In Ghahramani, Z., Welling, M., Cortes, C., Lawrence, N., and
  Weinberger, K. (eds.), \emph{Advances in Neural Information Processing
  Systems 27}, pp.\  1278--1286. Curran Associates, Inc., 2014.

\end{thebibliography}
\end{document}